\documentclass[pdflatex,sn-mathphys-num]{sn-jnl}

\usepackage{amssymb} 
\usepackage{pifont} 
\newcommand{\cmark}{\ding{51}} 
\newcommand{\xmark}{\ding{55}} 
\usepackage{arydshln}
\usepackage{graphicx}%
\usepackage{multirow}%
\usepackage{amsmath,amssymb,amsfonts}%
\usepackage{amsthm}%
\usepackage{mathrsfs}%
\usepackage[title]{appendix}%
\usepackage{xcolor}%
\usepackage{textcomp}%
\usepackage{manyfoot}%
\usepackage{booktabs}%
\usepackage{algorithm}%
\usepackage{algorithmicx}%
\usepackage{algpseudocode}%
\usepackage{listings}%
\usepackage[T1]{fontenc}
\usepackage[utf8]{inputenc}
\usepackage{babel}
\usepackage[font=small,labelfont=bf,tableposition=top]{caption}
\usepackage{booktabs}
\usepackage{threeparttable}
\usepackage{lmodern}
\usepackage{mathrsfs}

\begin{document}

\title[Article Title]{MolNexTR: A Generalized Deep Learning Model for Molecular Image Recognition}


\author[1]{\fnm{Yufan} \sur{CHEN}}\email{ychenkv@connect.ust.hk}

\author[1]{\fnm{Ching Ting} \sur{LEUNG}}\email{ctleungaf@connect.ust.hk}

\author[3]{\fnm{Yong} \sur{HUANG}}\email{yonghuang@ust.hk}

\author[1,3]{\fnm{Jianwei} \sur{SUN}}\email{sunjw@ust.hk}

\author*[1,2]{\fnm{Hao} \sur{CHEN}}\email{jhc@cse.ust.hk}

\author*[1]{\fnm{Hanyu} \sur{GAO}}\email{hanyugao@ust.hk}


\affil[1]{Department of Chemical and Biological Engineering, Hong Kong University of Science and Technology, Hong Kong SAR, China}

\affil[2]{Department of Computer Science and Engineering, Hong Kong University of Science and Technology, Hong Kong SAR, China}

\affil[3]{Department of Chemistry, Hong Kong University of Science and Technology, Hong Kong SAR, China}



\abstract{In the field of chemical structure recognition, the task of converting molecular images into machine-readable data formats such as SMILES string stands as a significant challenge, primarily due to the varied drawing styles and conventions prevalent in chemical literature. 
To bridge this gap, we proposed MolNexTR, a novel image-to-graph deep learning model that collaborates to fuse the strengths of ConvNext, a powerful Convolutional Neural Network variant, and Vision-TRansformer. 
This integration facilitates a more detailed extraction of both local and global features from molecular images.
MolNexTR can predict atoms and bonds simultaneously and understand their layout rules. It also excels at flexibly integrating symbolic chemistry principles to discern chirality and decipher abbreviated structures.
We further incorporate a series of advanced algorithms, including an improved data augmentation module, an image contamination module, and a post-processing module for getting the final SMILES output. 
These modules cooperate to enhance the model's robustness to diverse styles of molecular images found in real literature.
In our test sets, MolNexTR has demonstrated superior performance, achieving an accuracy rate of 81-97\%, marking a significant advancement in the domain of molecular structure recognition.

Scientific contribution: MolNexTR is a novel image-to-graph model that incorporates a unique dual-stream encoder to extract complex molecular image features, and combines chemical rules to predict atoms and bonds while understanding atom and bond layout rules. 
In addition, it employs a series of novel augmentation algorithms to significantly enhance the robustness and performance of the model.
}




\maketitle
\section{Introduction}
\label{sec:introduction}
\begin{figure}[t]
\centering
\includegraphics[width=1\textwidth]{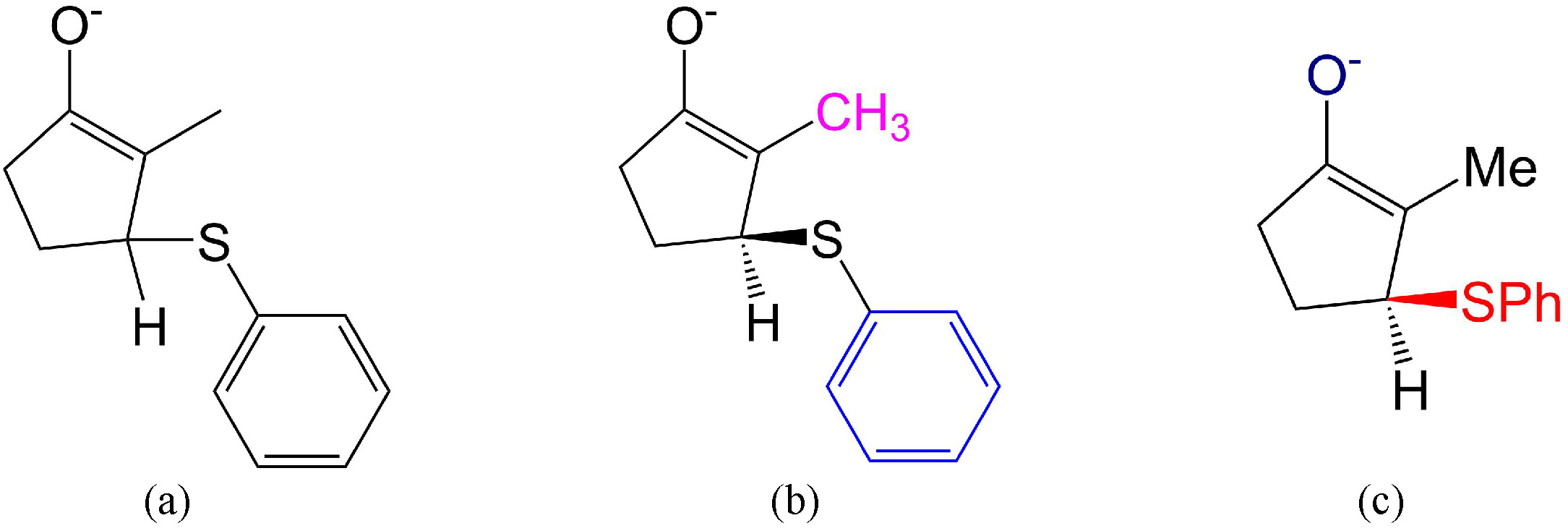}
\caption{The same molecule represented by three different styles. (a) depicts the full molecular structure and does not contain chirality information. (b) and (c) specify the chirality and use different abbreviations and colors.}
\label{example:1}
\end{figure}

In recent years, with the widespread development of deep neural networks, the performance of optical recognition tasks has significantly improved. 
However, the recognition of graphical or weakly structured information, such as molecular structure images, remains a challenging problem.
In chemical literatures, molecules are usually represented in the form of 2D images.
First, the drawing styles of molecules (atomic label fonts, bond drawing styles, etc.) are very diverse and not fully standardized among publishers.
Fig.~\ref{example:1} shows three different drawing styles of the same molecule. Second, molecules are often drawn as Markush structures. There are no general guidelines for the Markush structure, which leads to much variation in the Markush representation.
Third, there are instances where authors of chemical papers employ an artistic flair in representing chemical structures. 
Even experienced chemists can sometimes struggle to understand these more unusual representations.
For the reasons mentioned above, the existence of a wide variety of styles in molecular images makes the task of translating these images into machine-understandable molecular structures complex and challenging.
Previous work has demonstrated that it is possible to train convolutional neural network (CNN)-based or vision-transformer (ViT)-based models~\cite{staker2019molecular,rajan2020decimer,qian2023molscribe,rajan2021decimer,xu2022swinocsr,yoo2022image,clevert2021img2mol,image2smiles} to perform well in some specific styles. 
However, their robustness and generalization are not guaranteed across so many drawing styles. 
Moreover, it is impossible to guarantee that the training data contains all possible styles and patterns. 
Furthermore, ViT-based methods have limitations in translation invariance and local feature representation, and CNN-based methods have limitations in global feature representation. Both local and global features are crucial components for molecular structure recognition, since both the information of the atoms themselves and the information between the atoms are essential.
In the medical imaging field, some advanced  CNN-ViT hybrid architectures have been proposed to address this problem~\cite{cao2022swin,chen2021transunet,hatamizadeh2021swin,hatamizadeh2022unetr,oktay2018attention}. 
However, such methods have not yet emerged in the field of molecular structure recognition.

In this work, we aim to enhance the robustness and generalization of the molecular structure recognition model by enhancing its feature extraction ability and augmentation strategies, which can deal with any molecular images that may appear in the real literature.
We propose MolNexTR (Molecular convNext-TRansformer), which is defined as a graph generation model.
The model follows the encoder-decoder architecture, takes three-channel molecular images as input, outputs molecular graph structure prediction, and can be easily converted to machine-readable data formats such as simplified molecular-input line-entry system (SMILES)~\cite{weininger1988smiles}. 
The detailed steps are as follows. 
First, in the data pre-processing stage, we propose an image contamination augmentation strategy to simulate the interference molecular images (containing other words or other molecular fragments) from real literature. 
We also propose a series of data augmentation strategies to generate images with different drawing styles during training.
Second, in the encoder stage, we propose a novel combined CNN and ViT molecular image encoder network that is aware of both local atom information representation and long-range interatomic feature dependencies in the learning process.
Third, in the decoder stage, we combine the Pix2seq and Relationformer architecture~\cite{chen2021pix2seq,shit2022relationformer}, using an autoregressive decoder to predict atoms and their coordinates as a sequence and predict the chemical bonds between the atoms, composing the 2D molecular graph.
Finally, we include chemical knowledge as symbolic constraints to the model, such as determining the chirality of atoms from the predicted pattern, so that it can accurately recognize complex chemical molecules. 
An algorithm is used to resolve abbreviated functional groups commonly found in molecular images.
We also designed an error correction algorithm to improve the accuracy of the model for more complex abbreviated functional groups.

MolNexTR combines the advantages of deep learning model-based and chemical rule-based approaches. 
It combines the advantages of CNN and ViT for local information representation extraction and global information extraction, making it better able to deal with various styles of molecular images. 
It is also more robust to various disturbances of the image. 
The chemical rule-based approach allows the model to effectively enforce chemical constraints at inference time. 
This design allows MolNexTR to robustly identify local atoms and atomic bonds and make the most correct predictions using chemical rules.

Experiments show that MolNexTR outperforms prior image-to-SMILES or image-to-graph-based models on both in-domain and out-of-domain images and achieves strong recognition accuracy (81-97\%) on five public benchmarks. 
Meanwhile, on a previously constructed new benchmark containing molecular images from journal publications, MolNexTR significantly outperforms the existing methods, which illustrates the excellent generalization of MolNexTR. 
In addition, MolNexTR is more robust than existing methods in handling other disturbances such as low-quality images and input perturbations.
The main contributions of this paper are: 
\begin{itemize} 
\item We propose a novel combined CNN and ViT molecular image encoder network that leverages both local atom information and long-range intermolecular dependencies in the learning process. 
\item We propose a transformer-based decoder network with two predictive tasks that enable the model to not only identify the components of molecules but also understand the complex layout rules between them.
\item We propose a series of image and molecular augmentation algorithms that increase the generalization and robustness of the model.
\item Our proposed MolNexTR uniquely integrates deep learning model-based techniques with chemical rule-based methods in the post-processing part. This enables the effective processing of molecular images and the enforcement of chemical constraints during inference.
\item Our proposed MolNexTR demonstrates exceptional performance on multiple challenging datasets including Indigo, ChemDraw, RDKit, CLEF, UOB, JPO, USPTO, Staker and ACS.
\end{itemize}
\section{Related work}
\label{sec:Related work}
Molecular structure identification is a key task in the fields of cheminformatics and computational chemistry with origins dating back to the last century~\cite{rajan2020review}. It is also known as optical chemical structure recognition (OCSR).
Initially, scholars relied on conventional image-processing methodologies and rule-based systems to discern and analyze chemical structures. 
These basic systems~\cite{filippov2009optical,international1975abstracts,algorri2007reconstruction,casey1993optical,valko2009clide,frasconi2014markov,ibison1993chemical,mcdaniel1992kekule,park2009automated,sadawi2012chemical} depended on image binarization, line smoothing, vectorization techniques, and optical character recognition (OCR) to identify atoms and bonds within molecular images. Nonetheless, these frameworks typically needed manual rules and heuristic approaches created by chemists to tackle diverse scenarios, constraining their widespread applicability and accuracy to some extent. As time progressed, the open-source community contributed several tools~\cite{smolov2011imago,filippov2009optical}, addressing newfound challenges in molecular recognition, such as bridge bond identification. 
Though these systems demonstrated commendable performance on patent images, their accuracy decreased when dealing with the multifaceted images of journal articles.

With the rapid evolution of deep learning and neural network technologies in recent years, researchers have embarked on exploring novel methodologies for molecular structure recognition. 
Notably, by employing a convolutional neural network and a recurrent neural network (RNN), researchers devised a novel image-to-SMILES string generation model~\cite{staker2019molecular}, facilitating automated recognition of molecular structures. 
After that, a plethora of neural network-based model architectures, such as the Inception network~\cite{rajan2020decimer},  ResNet~\cite{image2smiles}, Transformer~\cite{rajan2021decimer}, Swin Transformer~\cite{xu2022swinocsr,qian2023molscribe} and pretrained-decoder~\cite{clevert2021img2mol}, have been proposed to augment the recognition accuracy and robustness of these systems.

Despite the success of neural network-based approaches, several challenges persist. 
Prevailing neural models often struggle with minuscule image variations and noise, and encounter difficulties when grappling with stereochemistry and abbreviated structures issues. 
Concurrently, since these models predominantly operate on SMILES strings rather than explicitly recognizing atoms and bonds, integrating chemical rules and constraints within them proves challenging.

To mitigate these issues, novel systems like ChemGrapher~\cite{oldenhof2020chemgrapher} and MolMiner~\cite{xu2022molminer} have adopted distinctive approaches. They train separate modules to detect atoms, bonds, and texts, followed by heuristic-based graph construction, thus allowing the integration of chemical constraints during the construction phase. 
Qian et al. and Yoo et al. have adopted an end-to-end model to generate molecular graphs~\cite{qian2023molscribe,yoo2022image}.
This approach eschews reliance on heuristic methods for connecting local predictions, simplifying the model architecture, and paving new avenues for the future evolution of molecular structure recognition technology. 

Our model followed this streamlined architecture, further enhancing model performance while bolstering robustness and generalization capabilities.
\section{Methods}
\label{sec:method}

\subsection{Problem Definition}
Molecular structure recognition aims to transfer images of individual molecules into their corresponding molecular structures with information of atoms and bonds.
In our work, we define this problem as a graph sequence formulation. 
Given an image $I$ of a single molecular, we transfer it into graph sequence $S^G=(S^M,S^N)$ that represents molecular structures, where $S^M=[M_1,M_2,...,M_n]$ is the atoms sequence and $S^N=[N_1,N_2,...,N_n]$ is the bond sets sequence. 
Specifically, $M_n = (l_n, x_n, y_n)$ is the definition of each atom, $l_n$ is the corresponding SMILES string of the atom and $(x_n, y_n)$ is the 2D coordinates of the atom in the image. 
$N_n=[m_{n,1}T,m_{n,2}T,...,m_{n,n}T]$ is the set of bonds of each atom, $m_{n,1}T$ means the $n$-th atom is connected to the $1st$ atom with bond $T$, where $T$ is the set of bond types that represents a different type of bonds (e.g., single, double, solid wedge, dashed wedge, or none).
Finally, it can be easily converted into molecular graphs and standard data formats.

\subsection{Data augmentation}
We first design a series of data augmentation methods for both images and molecules before the train data is fed into the model.
The train data construction details can be found in Sec~\ref{sec:dataset}. 
The data augmentation include several augmentation methods and an image contamination algorithm, which allowed the training data to cover a wide variety of styles and chemical patterns in the real literature, as well as possible interference.
\subsubsection{Rendering Augmentation}
In the process of rendering molecular images, we randomly render changes in the style and geometry of the molecules. It provides coverage for different drawing styles, improving the robustness of the model. 

To ensure the variety of rendering styles, we randomly selected one of RDKit~\cite{landrum2013rdkit} and Indigo~\cite{pavlov2011indigo} as the rendering tools and modified the source code to obtain the atomic coordinates. We randomly use options in both tools such as various relative thicknesses, various bond-line widths, various font families and sizes various distances between lines in double and triple bonds and different label modes, and visible implicit hydrogens.

\subsubsection{Image Augmentation}
After completing Render Augmentation, we further applied some perturbations to the molecular images.  We applied the following  image augmentation methods:
\begin{itemize} 
\item Rotate by a random angle,

\item Crop each side of the image,

\item Pad one side of the image, 

\item Blur the image,

\item Downscale the image,

\item Randomly compress, and enlarge Images,

\item Add Gaussian noise to the image,

\item Add salt-and-pepper noise to the image.
\end{itemize}

Similar to rendering-based augmentation, this step also aims to increase the style diversity of the training data and improve the robustness of the model.

\subsubsection{Molecular Augmentation}
\begin{figure}[t]
\centering
\includegraphics[width=1\columnwidth]{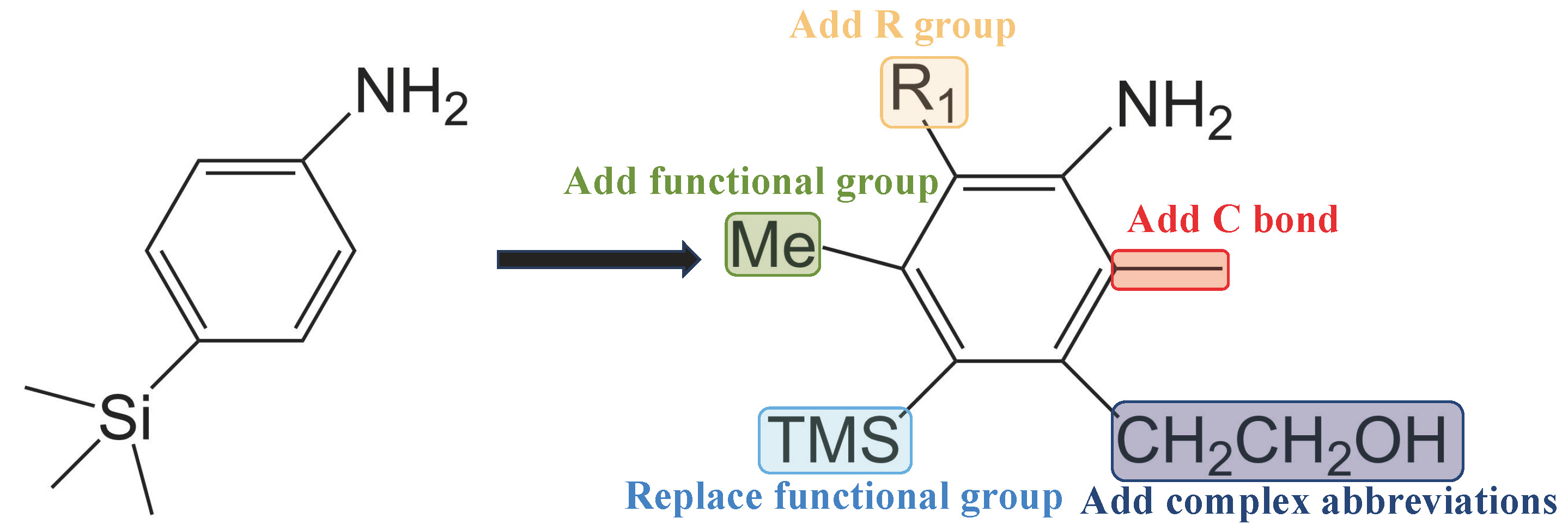}
\caption{The illustration of the molecular augmentation. The Molecular Augmentation consists of four main actions: 1) replace functional group, 2) add complex abbreviations, 3) add C bond, 4) add R-group.}
\label{MA}
\end{figure}
In most chemistry literature, authors map molecules with various functional groups abbreviations and R-group substituents. However, when rendering images using RDKit or Indigo, functional groups or R-groups are never included. 
To generate molecules with such abbreviations and substituents, we created a list of more than 100 common functional groups with their corresponding abbreviations. Augmenting algorithms randomly replace functional groups with abbreviations in molecules to generate augmented datasets. Specifically, in the process of molecular augmentation, if a functional group exists in a molecule, we randomly replace this functional group with the corresponding abbreviation in the list according to a given probability. The original structure branch is removed from the molecular graph. 
To provide a variety of R-groups, we also have a list of R-group labels and randomly add them to the molecule. We also randomly added noise keys to simulate possible interference in real images.
Furthermore, another possible style of abbreviations is chain abbreviations such as CH3CH2NH2, which are longer and more complex than the abbreviations in the list and are impossible to list completely, but converting them to SMILES is easy as long as the right characters are identified. Therefore, for the model to generalize better to those not covered in the list, we also have a collection of chained abbreviation components (e.g., CH3, CH2, NH2, OH) that are randomly combined to form a complex abbreviation. We believe that this method can improve the model capability of OCR so that the model can recognize such unseen abbreviations and will not ignore abbreviations whose length and complexity are far beyond the known abbreviations in the list.
The process of our molecular augmentation is illustrated in Fig.~\ref{MA}. 

\subsubsection{ Image Contamination Algorithm}

\begin{figure}[t]
\centering
\includegraphics[width=1\columnwidth]{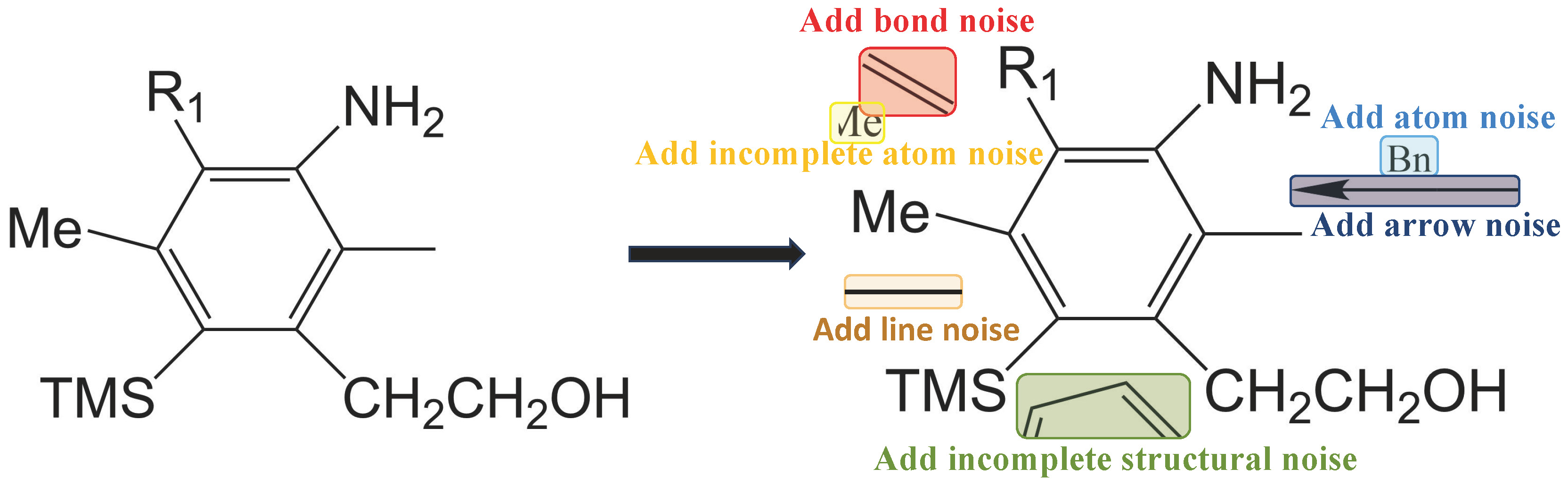}
\caption{The illustration of the image contamination algorithm. The Contamination Algorithm consists of six main actions which are 1) add atom noise, 2) add bond noise, 3) add incomplete structural noise, 4) add line noise, 5) add incomplete atom noise, and 6) add arrow noise.}
\label{ICA}
\end{figure}

When capturing images of molecules in real literature, they often appear as part of a chemical reaction, so the molecular images are often contaminated with other details, such as parts of other molecules, text, arrows, lines, and other elements.
Sometimes these objects are too close to the main molecule, or even cross or overlap with the main molecule, or they are inside the outline of the molecule and cannot be eliminated when the molecular image is captured. Experiments show that molecular predictions often fail because of these contaminations. 
However, the model should be robust to such contamination. 
To solve this problem, we propose an image contamination algorithm that simulates on typical pollution.
The illustration of our image contamination algorithm is in Fig.~\ref{ICA}.
Our algorithm randomly renders common pollution noise types:  1) atom noise, 2) bond noise, 3) incomplete structural noise, 4)line noise, 5) incomplete atom noise, and 6) arrow noise.
In order to prevent the pollution from overlapping with the main molecule or being too close to the main molecule, which leads to the instability of the model, when applying the algorithm, we first detect the effective pixels of the main molecule and then set a minimum distance from all the effective pixels to generate all the pollution outside this distance.

These data augmentations ensure that our model is trained with data that can cover a wide variety of style and chemical patterns as well as possible interference in real literature, thus better generalizing to practical usage scenarios.

\subsection{Network Overview}
\label{sec:netoverview}
\begin{figure*}[t]
\centering
\includegraphics[width=1\textwidth]{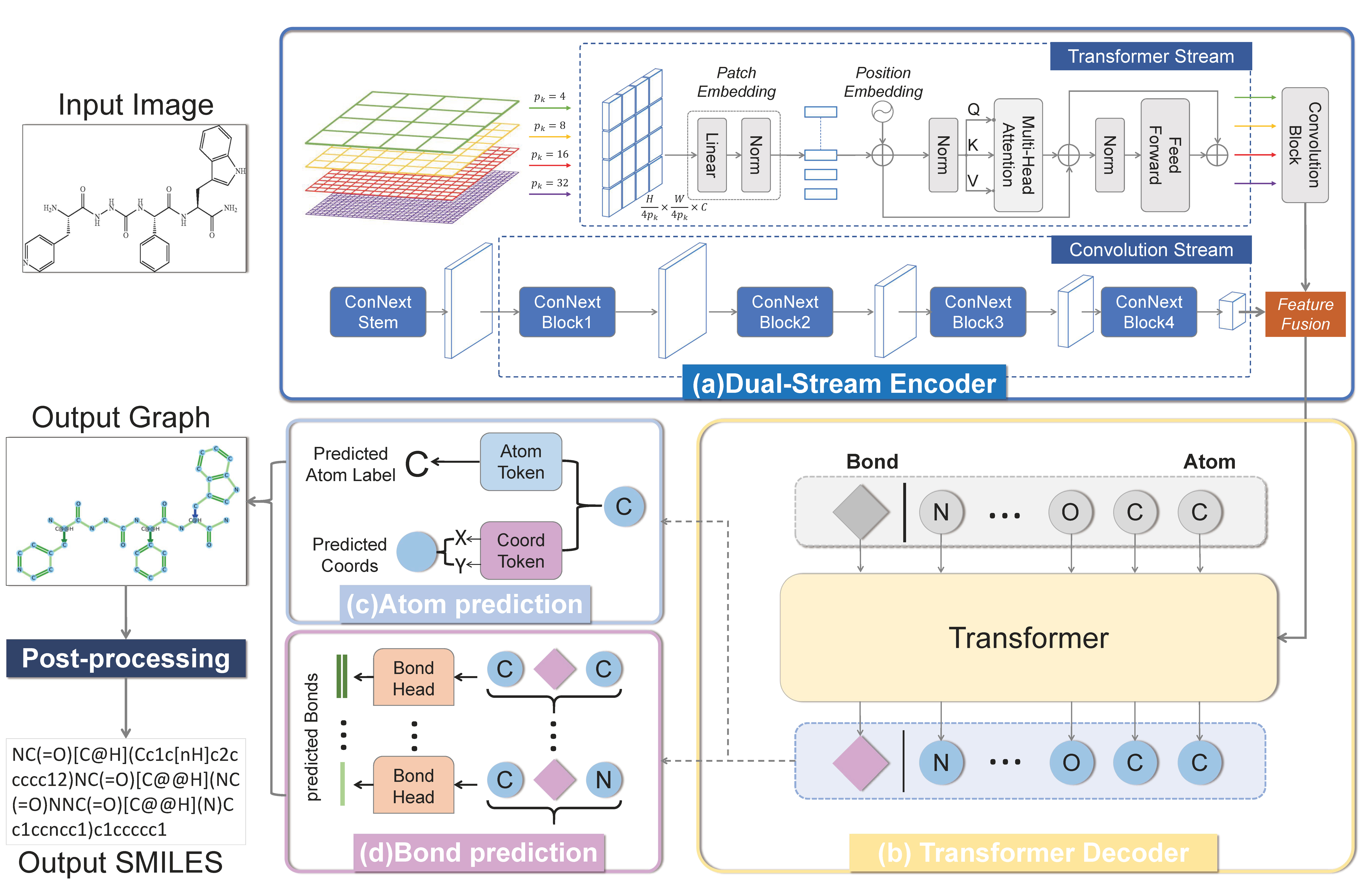}
\caption{Overview of our MolNexTR model. The molecular image is first encoded with the dual-stream encoder. Then a transformer-based decoder is applied for sequential atom and bond prediction. Finally, the post-processing modules ensure that the various molecular structures can be accurately converted to SMILES, SMART or MOLfile formats.}
\label{fig:1}
\end{figure*}

The Overview of our MolNexTR is illustrated in Fig.~\ref{fig:1}. It follows the encoder-decoder architecture and consists of three main components: 1) For the encoder network, we propose a dual-stream encoder (\textit{ref.}~Sec.~\ref{sec:encoder}), which concurrently combines a CNN and a vision transformer network to capture local feature dependencies and long-range feature dependencies, respectively. Such a design can avoid the problems of incomplete atom information (lack of local features) and incomplete relationships between atoms (lack of global features) as much as possible.; 
2) After the input image is encoded into a feature representation, we use a transformer-based decoder with a two-stage prediction (atom prediction and bond prediction) as the structure decoder network, so that our decoder can predict each atom and bond in the molecule, as well as their geometric arrangement. 
3) Afterward, we employ a post-processing module to ensure that the model can accurately and comprehensively reconstruct structures of various molecular styles. Finally, these structures can be easily converted into a SMILES string, a smiles arbitrary target specification (SMARTS) string, or a MOLfile~\cite{dalby1992description}.

\subsection{Dual-Stream Encoder Network}
\label{sec:encoder} 
\subsubsection{The CNN Stream}
The CNN stream is utilized to capture short-range feature dependencies of the given image. To this end, we choose ConvNext~\cite{liu2022convnet}, an efficient and powerful backbone that comprises of one convolutional stem and four ConvNext modules, where each module has a scale dimension of 4. As illustrated in the top of Fig.~\ref{fig:1}(a), ConvNext is a variant of the classical ResNet~\cite{he2016deep} that employs the split attention mechanism to capture multi-scale feature representations. Based on ConvNext, the convolutional backbone network can generate feature maps $F_c^k$ with a spatial resolution of $H/4 \times W/4$, $H/8 \times W/8$, $H/16 \times W/16$, and $H/32 \times W/32$, respectively. Unless stated otherwise, we follow the default architecture of the ConvNext backbone as described in the referenced paper.
\subsubsection{The ViTs Stream}
\label{sec:mst}
In addition to short-range feature dependencies, long-range feature dependencies between image patches also matter for molecular recognition. To this end, the long-range feature dependencies in different feature scales are captured by the ViTs stream~\cite{dosovitskiy2020vit}, which is composed of multiple parallel transformer blocks that receive feature patches under different scales as input. All transformer blocks share a similar structure, consisting of patch embedding layers and transformer encoding layers. As illustrated in the bottom of Fig. \ref{fig:1}(a), it shows the implementation process of ViT on the input feature map $\text{F}_1^c \in \mathbb{R}^{\frac{H}{4}\times \frac{W}{4}\times C}$. Firstly, the input feature map is divided into $\frac{HW}{16p^2}$ patches with size $p\times p$, and each patch is flattened into a vector $\mathbf{v}_n \in \mathbb{R}^{p^2\times C}$. In this work, four parallel transformer blocks are used, which receive feature patches of size $p=4,8,16,32$. Then, a linear projection layer is applied to each patch vector to obtain the patch embedding $\text{e}_n \in \mathbb{R}^{C}$. The patch embeddings, together with position embeddings, are then fed into the transformer encoding layers to obtain the output. Each encoding layer comprises a lightweight multi-head self-attention (MHSA) layer and a feed-forward network (FFN). The MHSA layer receives a truncated query $Q$, key $K$, and value $V$ as input and computes the attention score $\text{A} \in \mathbb{R}^{N\times N}$ as follows:
\begin{equation}
    \text{A} = \text{softmax}\left(\frac{QK^T}{\sqrt{d_k}}\right)V,
\end{equation}
where $N$ is the size of the patch number and $d_k$ is the dimension of the key. Empirically, truncating the feature vectors within a certain range does not diminish the model's recognition performance. Instead, it can significantly reduce the computational cost. The output of the MHSA layer is then fed into the FFN to obtain the output $\text{F}_t$:
\begin{equation}
    \text{F}_t = \text{FFN}(\text{A}),
\end{equation}
where $\text{FFN}$ is the feed-forward network with two $3\times3$ convolutional layers with the ReLU activation function. Finally, $\text{F}_t$ is reshaped into the same size as $\text{F}_c^1$ to obtain the output. The other transformer blocks are processed similarly. All outputs of the transformer blocks are concatenated along the channel dimension and fed into the convolutional layer to obtain the final output.

\subsection{Structure Decoder Network}
\label{sec: Structure edecoder}
\subsubsection{Transformer Decoder}
\label{sec: Transformer Decoder}
As Shown in Fig.~\ref{fig:1}(b), we chose the classic transformer decoder as our main decoder. Specifically, it has 6 Transformer blocks with 8 attention heads, a hidden
dimension of 256, and Sinusoidal position embedding.

During the decoding phase, we formulate the transformation from image to graph sequence as an autoregressive conditional probability generation:

\begin{equation}
    P(S^G\mid I) = P(S^M\mid I)P(S^N\mid S^M,I)
\end{equation}
Where $P(S^M\mid I)$ and $P(S^N\mid S^M,I)$ represent the atom prediction and bond prediction respectively. 

\subsubsection{Atom Prediction}
\label{sec: Atom prediction}
The atom prediction is crucial because it establishes the foundation of the molecular structure by identifying the types of atoms present and their coordinates, which is a precursor to understanding the molecule's geometry.
As Shown in Fig.~\ref{fig:1}(c), our model predicts atoms and their corresponding coordinates simultaneously by two heads. First we construct a sequence of specific tokens as the output:
\begin{equation}
    S^M=[M_1,M_2,...,M_n] = [l_1, x_1, y_1,l_2, x_2, y_2,...,l_n, x_n, y_n]
\end{equation}
where each atom $M_i$ is represented by three tokens $l_i, x_i, y_i$. $l_i$ is the SMILES string of the atom itself, including the element symbol, implicit hydrogen, charge, and all necessary information.  $x_i$ and $y_i$ represent the 2D coordinates of this atom in the original molecular image. Then the model generates the sequence autoregressively by a product of conditional probabilities:

\begin{equation}
    P(S^M\mid I) = \prod_{i=1}^{S^M} P(S^M_i\mid S^M_{<i},I)
\end{equation}

\subsubsection{Bond Prediction}
\label{sec: Bond prediction}
The bond prediction stage builds upon the atom prediction by determining how the identified atoms are connected. This step is fundamental to understanding the molecule's topology and its chemical properties.
Our model predicts the bonds between each pair of atoms as illustrated in Fig.~\ref{fig:1}(d). Similar to the atom prediction, we also construct a sequence as the output of bond prediction:
\begin{equation}
    S^N = [N_1, N_2, \ldots, N_n], \quad N_n = [m_{n,1}T, m_{n,2}T, \ldots, m_{n,n}T]
\end{equation}
Where $N_i$ represents the set of bonds between the $i$-$th$ atom and every other atom. It can also be viewed as an $n \times 1$ vector.
For each atom $M_i$, $N_i$ is the hidden state derived from the final token in the decoder's output. 
$m_{i,j}T$ denotes that the $i$-th atom is connected to the $j$-th atom by a type of bond $T$. $T = \{\text{"None"}, \text{"single"}, \text{"double"}, \text{"triple"}, \text{"aromatic"}, \text{"s.w"}, \text{"d.w"}\}$ which contains all types of bonds that may appear in the image, where "None" means no bond, "s.w" and "d.w" mean solid wedge and dashed wedge respectively.

For the bond prediction between each pair of atoms $M_i$ and $M_j$, we formulate it as a product of conditional probabilities in an autoregressive generation manner:
\begin{equation}
    P(S^N\mid S^M,I) = \prod_{i=1}^{n}\prod_{j=1}^{n} P(m_{i,j}T\mid S^M,I)
\end{equation}

The combination of these two predictive tasks within a multi-task learning framework ensures that the model not only identifies the components of the molecule but also understands the intricate relationships between them, leading to a comprehensive and accurate molecular graph. 
This graph is then post-processed to correct any inconsistencies and translated into a SMILES string.

\subsection{Post-processing Module}
\label{sec: Post-processing Module}
After the molecular graph is obtained, the final SMILES prediction is obtained through the utilization of a post-processing module. In the context of graph-to-SMILES translation, our focus lies on two crucial aspects: stereochemistry and abbreviations.

\subsubsection{Stereochemical discrimination module}
\label{sec: Stereochemistry}

The stereochemistry of organic molecules exerts a significant impact on their structure and properties.
However, recognizing stereochemistry poses challenges for neural models.
In SMILES notation, chirality is indicated as an atomic property, with "@" or "@@" following an atom label to describe the relative spatial arrangement of connected bonds, following their order of listing. There's no direct mapping between image patterns and the presence of "@" or "@@".
Furthermore, it becomes even more challenging to discriminate when there are complex functional groups connected to the chiral center.

The comprehension of stereochemistry necessitates the application of geometric reasoning in three-dimensional space and chemical knowledge, skills that are not well-suited for conventional two-dimensional neural networks.
We propose a stereochemical discrimination module that applies chemical rules to explicitly define stereochemistry using predicted atoms, coordinates, and bonds.

For a chirality prediction, we 1) first get atom labels and corresponding coordinates and every chiral center, 2) then identify bonds linked to each chiral center, 3) if a functional group exists around chiral center, expand its first two atoms, 4) and finally infer their relative order using predicted atom coordinates, and explicitly assign chirality types. 
This step can be easily accomplished using RDKit tools.
Therefore, our approach, which combines neural networks with traditional chemical rules, can more accurately recognize stereochemistry compared to using a standalone neural network.

\subsubsection{Abbreviation expansion and self-correction module}

Abbreviations are also a challenging part of molecular recognition. When describing molecular structures, abbreviations are commonly used to represent complex functional groups, such as "Ph" for phenyl, "Ac" for acetyl, and "Bn" for benzyl.
These abbreviations act as "superatoms" in molecular diagrams.

To deduce the complete molecular structure from these abbreviated forms, additional processing steps are required. 
Current methodologies, whether rule-based or machine learning-driven, typically involve creating a list of common abbreviations along with their corresponding functional group structures and SMILES, using this list to substitute the superatoms during model inference. 
However, due to the combinatorial nature of chemical abbreviations, relying solely on a list to cover all possible combinations is impractical. For instance, combinations such as "BnO" or "OBn" refer to the same structure but in a different order, or more random and longer abbreviations such as "NHCOOH", although considered hyperatomic, are almost impossible to enumerate exhaustively.

In practice, we also found that if the model makes minor mistakes in recognizing hyperatomic characters such as identifying "OTBMS" as "OTTBMS", it can lead to incorrect prediction of the entire molecular structure.

We propose an abbreviation expansion and self-correction module, which is more generalized and robust than previous methods.
Although we also have a list of common abbreviations and their corresponding SMILES, our approach can also be extended to unseen forms.
In an abbreviation expansion process, we first 1) check whether the superatom is in the list, and if so, we directly output the corresponding SMILES; 2) otherwise, we split the superatom symbol into atomic characters, such as expanding "O2CH3" into "OOCHHH". 3) After deriving the list of atoms, we greedily connect them based on valence bonds according to the SMILES formalism until their valence bonds are full, at which point the output is the final SMILES. 4) If SMILES cannot be successfully output, we will compare the original abbreviation symbol with every abbreviation in the list, find the abbreviation with the highest similarity, and replace it with the corresponding SMILES if the similarity is greater than the set threshold $\sigma = 0.8$.

For the prediction of a single molecule, the final SMILES is output after the stereochemistry of all chiral centers have been discriminated and all abbreviations have been expanded.
\section{Experiments}
\label{sec:exp}
\subsection{Dataset}
\label{sec:dataset}
\subsubsection{Train data}
Training data in a variety of styles can help improve the robustness of the model, so our training dataset contains both Synthetic and real data, which comes from two main sources.

\paragraph{PubChem.}
PubChem database~\cite{kim2016pubchem} contains about 100M molecules. We use this database as the synthetic training data. However, they are not all required, we estimate that we need about 1M structures.
We randomly select 1M molecules from the database and render their images using chemical tools.

\paragraph{USPTO.} Following previous work~\cite{qian2023molscribe}, we collect 0.68M examples from USPTO~\cite{marco2015uspto} which contain molecular images and structure labels. We use this dataset as the real training data. 
Coordinate annotations of the data are obtained by normalizing the relative coordinates available in MOLfiles according to the image size.
This dataset is noisy and contains a variety of styles, which is closer to the actual use of the model than the synthetic data.

\subsubsection{Test data} We evaluate our MolNexTR on six public realistic datasets, which are CLEF, UOB, JPO, USPTO, Staker, and ACS. The ACS dataset is a new dataset collected by~\cite{qian2023molscribe} with 331 molecular images taken from ACS publications. Compared to other datasets, it is more diverse in terms of drawing styles and the use of abbreviations.
Furthermore, we created three synthetic datasets, generated by indigo, RDKit, and ChemiDraw, respectively. They were rendered from the same dataset of 5,719 molecules.
Table~\ref{tab:dataset} presents the details of the test sets we used.
\begin{table}[ht]
\caption{Summary of the test datasets}
\label{tab:dataset}

\centering
\renewcommand\arraystretch{1.5}
\begin{tabular}{ccccc}
\toprule
\textbf{Dataset} & \textbf{Type} & \textbf{Total number of images} & \textbf{abbreviations} & \textbf{chirality} \\
\midrule
Indigo & synthetic & 5,719 & \xmark & \cmark \\
ChemDraw & synthetic & 5,719 & \xmark & \cmark\\
RDKit & synthetic & 5,719 & \xmark & \cmark\\
\midrule
CLEF & real & 992 & \cmark & \cmark \\
UOB & real  & 5,740 & \cmark & \xmark\\
JPO & real  & 450 & \xmark & \xmark\\
USPTO & real  & 5,719 & \cmark & \cmark\\
Staker & real  & 50,000 & \cmark & \cmark \\
ACS & real & 331 & \cmark & \cmark \\
\bottomrule
\end{tabular}
\label{tab:chiral_datasets}
\footnotetext{\cmark~means inclusion, \xmark~means no inclusion.}
\end{table}

\subsection{Evaluation Metrics}
\label{sec:Metrics}

Following~\cite{staker2019molecular, clevert2021img2mol, yoo2022image}, the SMILES sequence exact matching accuracy is used as the primary accuracy evaluation metric. 
Specifically, we first transform both the final predicted SMILES and the ground truth into canonical SMILES, a unique molecular representation, and then calculate their exact sequence matching accuracy. 
For R-groups, we use (*) instead of R, and for other R-groups such as R1, we use (1*) instead.
For stereochemistry, we only consider matching tetrahedral chirality and ignore other forms of stereoisomerism, as the information is usually not present in the ground truth.
\subsection{Implementation Details}
\label{sec:implementation}
We optimized our model using the ADAM optimizer with a maximum learning rate of 3e-4 and a linear warmup for 5\% steps. The default batch size was set to 256 with the image size of 384$\times$384.
The CNN stream encoder was initialized with the pre-trained weights of ConvNext on ImageNet and then fine-tuned for 40 epochs on 10 NVIDIA RTX 3090 GPUs. 
The decoder was a 6-layer Transformer~\cite{vaswani2017attention} with 8 attention heads, a hidden dimension of 256, and sinusoidal positional encoding.  The dropout probability was set to 0.1.


\section{Results and Discussion}

\subsection{Comparison with Current Methods}
\begin{table}[t]
  \centering
  \caption{Comparison of our model's results with current methods across various test sets}
  \label{tab:sota}
  \renewcommand\arraystretch{1.2}
  \setlength{\tabcolsep}{2pt}
  \begin{tabular}{@{}ccccccccccc@{}}
    \toprule
    \multirow{2}{*}{Base} & \multirow{2}{*}{Methods} & \multicolumn{3}{c}{Synthetic} & \multicolumn{6}{c}{Realistic} \\
    \cmidrule(lr){3-5} \cmidrule(lr){6-11} 
    & & Indigo & ChemDraw & RDKit & CLEF & UOB & JPO & USPTO & Staker & ACS \\
    \midrule
    \multirow{2}{*}{Rule-based} & MoIVec~\cite{peryea2019molvec} & 95.4 & 87.9 & 88.7 & 82.8 & 80.6 & 67.8 & 88.4 & 0.8 & 47.4  \\
    & OSRA~\cite{filippov2009optical} & 95.0 & 87.3 & 88.2 & 84.6 & 78.5 & 55.3 & 87.4 & 0.0 & 55.3  \\
    \midrule
    \multirow{8}{*}{Deep learning-based} & MSE-DUDL~\cite{staker2019molecular} & --- & --- & --- & --- & --- & --- & --- & 77.0 & ---  \\
    & ChemGrapher~\cite{oldenhof2020chemgrapher} & --- & --- & --- & --- & 70.6 & --- & --- & --- & ---  \\
    & Image2Graph~\cite{yoo2022image} & --- & --- & --- & 51.7 & 82.9 & 50.3 & 55.1 & --- & ---  \\
    & SwinOCSR~\cite{xu2022swinocsr} & 74.0 & 79.6 & 77.3 & 30.0 & 44.9 & 13.8 & 27.9 & --- & 27.5  \\
    & Img2Mol~\cite{clevert2021img2mol} & 58.9 & 46.4 & 44.2 & 18.3 & 68.7 & 16.4 & 26.3 & 17.0 & 23.0  \\
    & DECIMER~\cite{rajan2020decimer} & 69.6 & 86.1 & 82.3 & 62.7 & \underline{88.2} &55.2& 41.1 & 40.8 & 46.5  \\
    & MolScribe~\cite{qian2023molscribe} & \underline{97.5} & \underline{93.8} & \underline{94.6} & \underline{88.3} & 87.9 & \underline{77.7} & \underline{92.6} & \underline{86.9} & \underline{71.9}  \\
    & Ours & \textbf{97.8} & \textbf{95.1} & \textbf{96.4} & \textbf{90.4} & \textbf{88.5} & \textbf{82.1} & \textbf{93.8} & \textbf{88.3} & \textbf{81.9} \\
    \bottomrule
  \end{tabular}
  \footnotetext{Scores are in overall SMILES sequence exact matching accuracy (\%). "---" denotes that the results are unavailable. Bold represents the best performance and underline represents the second-best performance}
\end{table}

\begin{table}[t]
  \centering
  \caption{Comparison of our model’s results with current methods across various perturbed test sets}
  \label{tab:pertu}
  \renewcommand\arraystretch{1.2}
  \setlength{\tabcolsep}{2pt}
  \begin{tabular}{@{}cccccccccccccc@{}}
    \toprule
    \multirow{2}{*}{Base} & \multirow{2}{*}{Methods} & \multicolumn{6}{c}{Perturbed by img transform.} & \multicolumn{6}{c}{Perturbed by curved arrows.} \\
    \cmidrule(lr){3-8} \cmidrule(lr){9-14}
    &  & CLEF & UOB & JPO & USPTO & Staker & ACS & CLEF & UOB & JPO & USPTO & Staker & ACS \\
    \midrule
\multirow{2}{*}{Rule-based} & MoIVec~\cite{peryea2019molvec} & 43.7 & 74.5 & 22.5 & 29.7 & 5.0 & 10.6 & 40.8 & 72.6 & 20.6 & 22.9 & 4.1 & 5.3 \\
& OSRA~\cite{filippov2009optical} & 11.5 & 68.3 & 10.8 & 4.0 & 4.6 & 12.3 & --- & --- & --- & --- & --- & --- \\
\midrule
\multirow{5}{*}{Deep learning-based} & SwinOCSR~\cite{xu2022swinocsr} & 32.2 & --- & --- & --- & --- & --- & 31.5 & --- & --- & --- & --- & ---\\
& Img2Mol~\cite{clevert2021img2mol} & 21.1 & 74.9 & 8.9 & 29.7 & 51.7 & 8.6 & ---& --- & --- & --- & --- \\
& DECIMER~\cite{rajan2020decimer} & 70.6 & \underline{87.3} & 34.1 & 46.4 & 47.9 & 20.1 & 59.8 & --- & 38.5 & --- & --- & 42.4 \\
& MolScribe~\cite{qian2023molscribe} & \underline{90.4} & 86.7 & \underline{52.0} & \underline{92.5} & \underline{65.0} & \underline{53.1} & \underline{87.1} & \underline{81.6} &\underline{56.7} & \underline{88.9} & \underline{73.0} & \underline{62.8} \\
& Ours & \textbf{90.4} & \textbf{88.4} & \textbf{71.9} & \textbf{93.2} & \textbf{78.9} & \textbf{68.8} & \textbf{89.5} & \textbf{84.3} & \textbf{67.7} & \textbf{91.2} & \textbf{73.2} & \textbf{70.8} \\
\bottomrule
  \end{tabular}
    \footnotetext{Scores are in overall SMILES sequence exact matching accuracy (\%). "---" denotes that the results are unavailable. Bold represents the best performance and underline represents the second-best performance}
\end{table}

\begin{figure*}[t]
\centering
\includegraphics[width=1\textwidth]{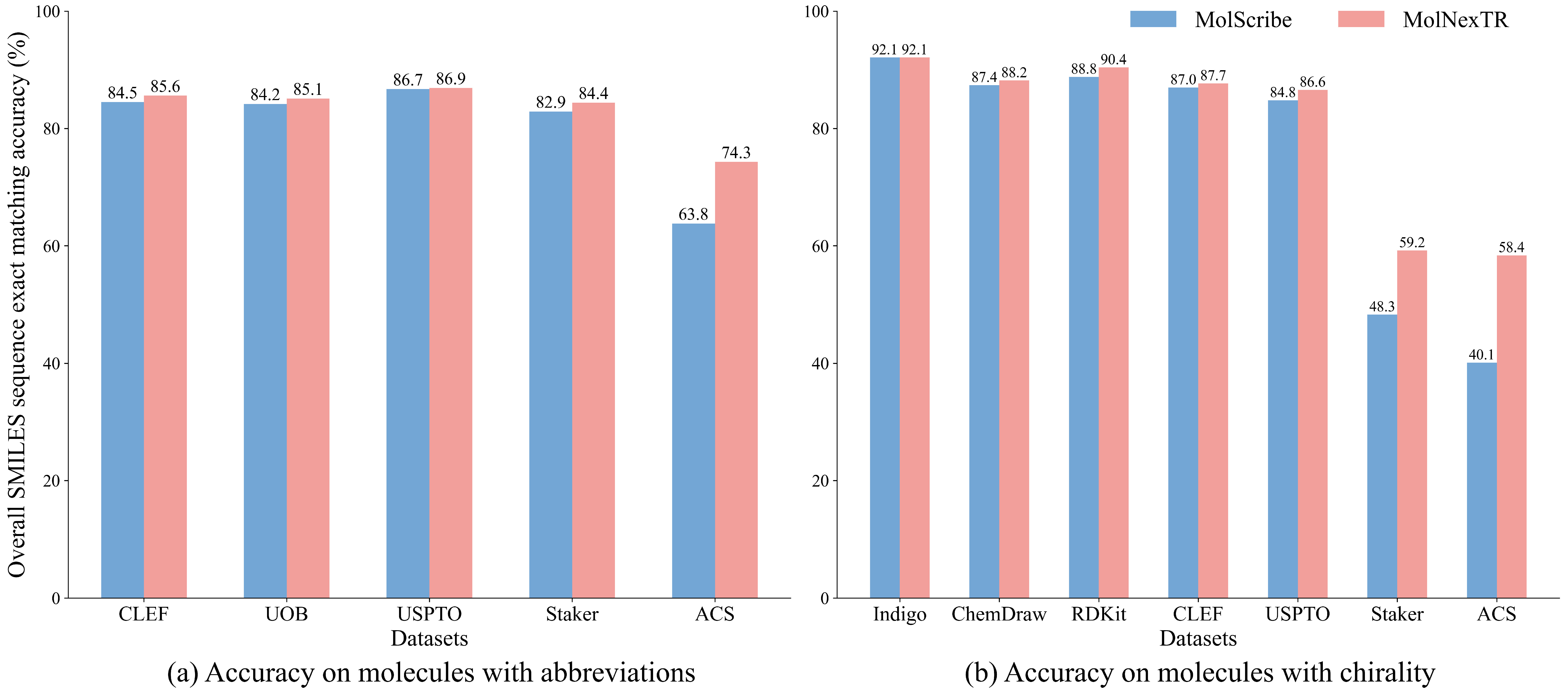}
\caption{Comparison of our model’s results with MolScribe on molecules with chirality or abbreviations.}                          
\label{fig:2}
\end{figure*}
We compare our MolNexTR with the current methods including rule-based OSRA (Version 2.1.3)~\cite{filippov2009optical} and MolVec (Version 0.9.8)~\cite{peryea2019molvec}, and deep learning-based models Img2Mol~\cite{clevert2021img2mol}, DECIMER (Version 2.1.0)~\cite{rajan2021decimer}, SwinOCSR~\cite{xu2022swinocsr}, MSE-DUDL~\cite{staker2019molecular}, ChemGrapher~\cite{oldenhof2020chemgrapher}, Image2Graph~\cite{yoo2022image}, and MolScribe (The best checkpoint)~\cite{qian2023molscribe}. We directly use the results reported in previous works.
The results are shown in Table~\ref{tab:sota}.
The rules-based systems MolVec and OSRA achieve good performance on CLEF, UOB, and USPTO, but decline on JPO due to the Japanese characters involved. The performance on ACS decreases more due to the extremely diverse drawing styles of this dataset. However, they dropped significantly on Staker, possibly due to the low resolution of this dataset.
Among deep learning methods, we can observe that for synthetic datasets, MolNexTR achieves 97.8\% in Indigo, 95.1\% in ChemDraw and 96.4\% in RDKit, which outperforms the second-best method by 0.3\%, 1.3\% and 1.8\% respectively.
On realistic datasets, our MolNexTR achieves 90.2\% in CLEF, 88.5\% in UOB, 82.1\% in JPO, 93.8\% in USPTO, and 88.3\% in Staker, respectively, which outperforms the second-best method by 1.1\%, 0.3\%, 4.4\%, 1.2\% and 1.4\%, respectively, validating the superior performance of our innovative model architecture. 
Notably, our MolNexTR achieves 81.9\% in ACS, which outperforms the second-best method by 10.0\%, which is a significant improvement. 
Compared with other datasets, the ACS has more diverse image styles and contains much more contamination.
This proves that our series of data augmentation methods solve these problems well and improve the generalization and robustness of the model.

We also evaluate the model on perturbed datasets with some image transform following the setup of Clevert et al., the results are shown on the left of Table~\ref{tab:pertu}.  MolNexTR performs better than the current methods, and the accuracy decays less. 
We further construct another perturbed dataset with curved arrows on it to simulate the molecules in the mechanism images, which is also a molecular pattern commonly found in the real literature, the results are shown on the right of Table~\ref{tab:pertu}. MolNexTR can better recognize such molecular patterns without being affected by the arrows in the molecules though we haven't included such molecules in the training process. These results further demonstrate that our model has excellent robustness to deal with disturbances.

We further compare the performance on molecules with chirality and abbreviation between MolNexTR and the MolScribe, which performs the best on multiple test datasets on average among current methods. The results are shown in Fig.~\ref{fig:2}.  MolNexTR shows better performance, especially on datasets with more diverse drawing styles. This is because we consider the presence of functional groups when judging chirality, and obtain the information of functional groups before judging chirality, which is not considered by MolScribe. When expanding abbreviations, the self-correction mechanism can improve the accuracy of judgment. On the other hand, our data augmentation method enables the model to have a better OCR ability.

\begin{table}[t]
\caption{Ablation study results on the superiority of the dual-stream encoder}
\label{tab:dual-stream encoder}
\centering
\renewcommand\arraystretch{1.5}
\setlength{\tabcolsep}{7pt} %
\begin{tabular}{ c c c c c c c} 
\toprule
CNN & Single ViT & Multiple parallel ViTs & Indigo & ChemDraw & CLEF & ACS \\
\midrule
\cmark & \xmark & \xmark & 96.7 & 93.8 & 88.8 & 80.3\\  
\cmark & \cmark & \xmark  & 97.0$_{{+0.3}}$ & 94.5$_{{+0.7}}$ & 89.4$_{{+0.6}}$ & 80.8$_{{+0.5}}$ \\  
\cmark  & \xmark & \cmark & 97.8$^*_{{+0.8}}$ & 95.1$_{{+0.6}}$ & 90.2$^*_{{+0.8}}$  & 81.9$^*_{{+1.1}}$ \\
\bottomrule
\end{tabular}
\footnotetext{\cmark~means inclusion, \xmark~means no inclusion. * means the component achieves significant performance improvement over 0.8\% with p $<$ 0.05 via paired t-test.}
\end{table}
\begin{table}[t]
\caption{Ablation study results on the effectiveness of the data augmentation algorithm}
\label{tab:dataag}
\centering
\renewcommand\arraystretch{1.5}
\setlength{\tabcolsep}{2.5pt}
\begin{tabular}{ cccc  cccc } 
\toprule
Render Aug. & Image Aug. & Mol Aug. & Img Contam Alg. & Indigo & ChemDraw & CLEF & ACS \\
\midrule
\xmark & \xmark & \xmark & \xmark & 92.0 & 80.6 & 75.4 & 60.8\\  
\cmark & \xmark & \xmark & \xmark  & 93.8$_{+1.8}$ & 85.3$^*_{+4.7}$ & 79.4$^*_{+4.0}$ & 65.0$^*_{+4.2}$ \\  
\cmark & \cmark  & \xmark & \xmark & 95.4$_{+1.6}$ & 88.8$^*_{+3.5}$ & 82.0$^*_{+2.6}$  & 68.9$^*_{+3.9}$ \\
\cmark & \cmark  & \cmark & \xmark & 96.8$_{+1.4}$ & 93.6$^*_{+5.8}$ & 88.9$^*_{+6.9}$  & 75.1$^*_{+6.2}$ \\
\cmark & \cmark  & \cmark & \cmark & 97.8$_{+1.0}$ & 95.1$_{+1.5}$ & 90.2$_{+1.3}$  & 81.9$^*_{+6.7}$ \\
\bottomrule
\end{tabular}
\footnotetext{\cmark~means inclusion, \xmark~means no inclusion. * means the component achieves significant performance improvement over 2\% with p $<$ 0.05 via paired t-test.}
\end{table}
\subsection{Ablation Study}

We conduct an ablation study to investigate the superiority of the dual-stream encoder and the effectiveness of each component of the data augmentation algorithm in MolNexTR. All ablation studies are conducted on Indigo, ChemDraw, CLEF, and ACS.

\subsubsection{Superiority of dual-stream encoder} 

We conduct ablation studies to explore the effectiveness of each component in the encoder in MolNexTR. In Table~\ref{tab:dual-stream encoder}, we compare the performance of MolNexTR variants on four datasets: 1) CNN, only the convolution stream; 2) $+$Single ViT, single patches ViT; 3)$+$Multiple parallel ViTs, the final dual-stream encoder with convolution and Transformer. All the components consistently boost the performance by 0.3\% and 0.8\% in Indigo, 0.7\% and 0.6\% in ChemDraw, 0.6\% and 0.8\% in CLEF, 0.5\% and 1.1\% in ACS, respectively. These findings indicate that the evaluated MolNexTR variants can effectively enhance the molecular prediction performance of the baseline model. Furthermore, our results show that adding multiple parallel ViTs on top of a CNN model leads to improved performance compared to CNN alone or adding single patches ViT. This finding suggests that combining different models can lead to better performance than using a single model.


\subsubsection{Effectiveness of data augmentation algorithm} 
\begin{figure*}[t]
\centering
\includegraphics[width=1\textwidth]{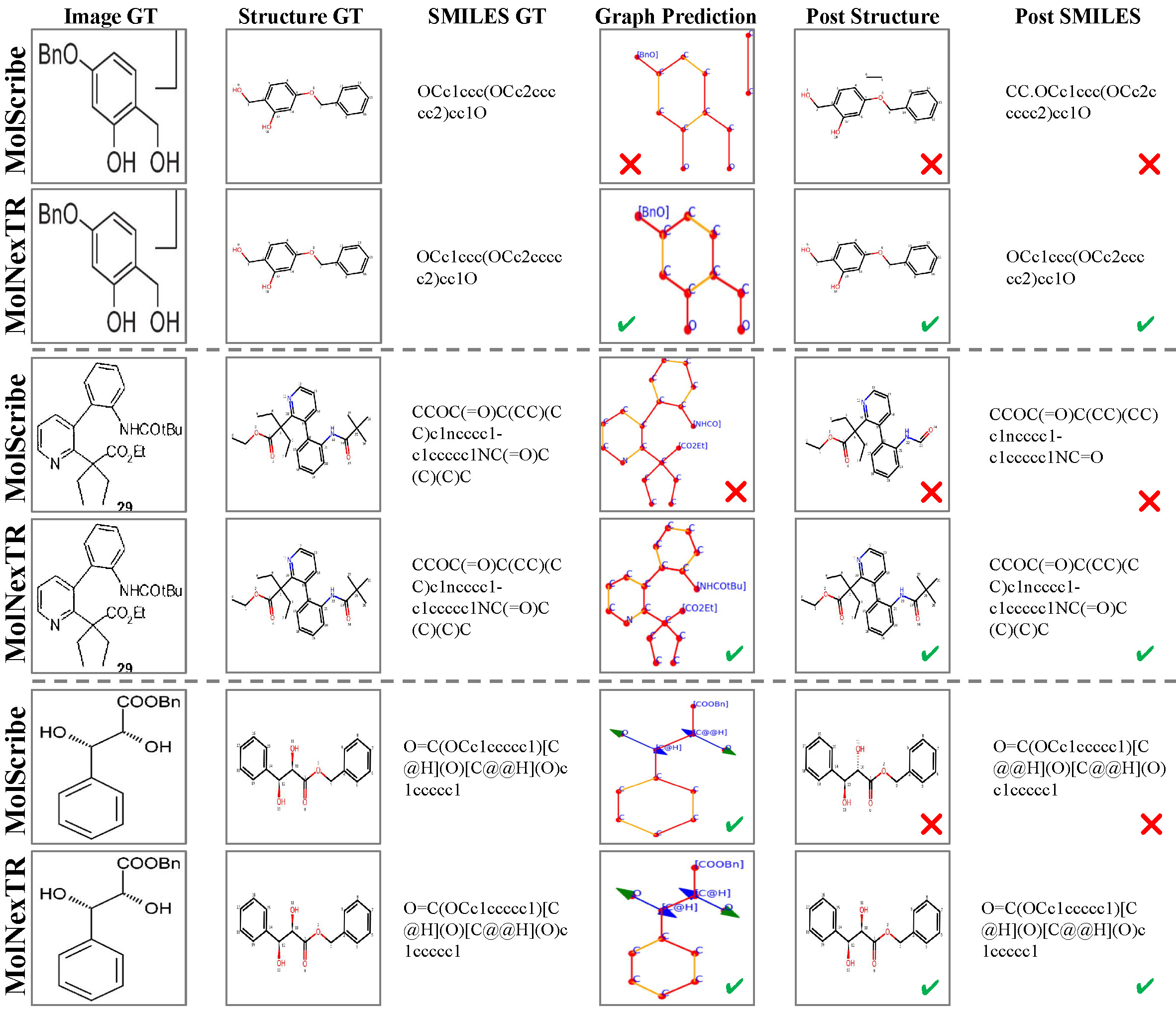}
\caption{Visualization of our model on the ACS dataset compared to MolScribe.}                          
\label{fig:vs1}
\end{figure*}
In this study, we conducted an experimental analysis to explore the effectiveness of each component in the data augmentation algorithm. The evaluated variants include $+$Render Augmentation, $+$Image Augmentation, $+$Molecular Augmentation, and $+$Image Contaminate Algorithm. Our results, presented in Table ~\ref{tab:dataag}, demonstrate that each of these data augmentations can significantly improve the baseline performance.
Specifically, when applied to the MolNexTR, $+$Render Aug. (Render Augmentation), $+$Image Aug. (Image Augmentation), $+$Mol Aug. (Molcular Augmentation), and $+$Img Contam Alg. (Image Contamination Algorithm) led to performance gains of 1.8\%, 1.6\%, 1.4\% and 1.0\% in Indigo, 4.7\%, 3.5\%, 5.8\% and 1.5\% in ChemDraw, 4.0\%, 2.6\%, 6.9\% and 1.3\% in CLEF, 4.2\%, 3.9\%, 6.2\% and 6.7\% in ACS, respectively. 
We can observe that for the in-domain dataset Indigo(which uses the same tools as the training data), the data augmentation method does not improve the accuracy much. However, once applied to the out-of-domain dataset, most components of the data augmentation can greatly improve the prediction accuracy, especially in the ACS dataset. Although the application of the image contamination algorithm has only slight improvement on indigo, ChemDraw, and CLEF, it has great improvement on ACS datasets with more diverse styles and more pollution. It can be said that ACS data sets are more closely related to real-world situations, so it is still necessary to apply the image contaminate algorithm. The above demonstrates the effectiveness of our data augmentation methods. They improve the robustness and generalization of the model that can be better applied in the real world.

\subsection{Qualitative Results}
\subsubsection{Visual Comparison with Current Methods}

\begin{figure*}[t]
\centering
\includegraphics[width=1\textwidth]{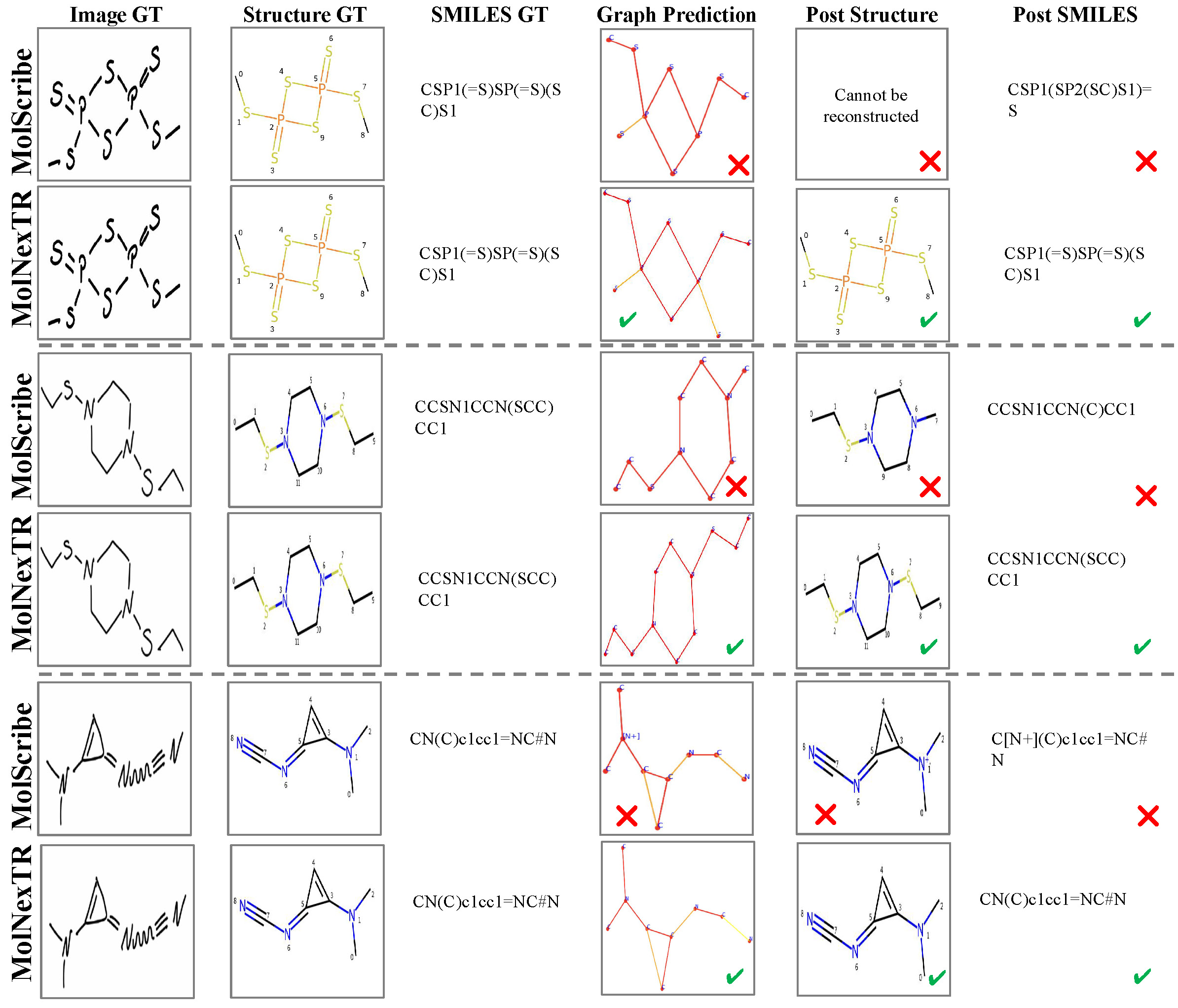}
\caption{Visualization of our model on some hand-drawn molecule images compared to MolScribe.}                          
\label{fig:vs3}
\end{figure*}
\begin{figure*}[t]
\centering
\includegraphics[width=1\textwidth]{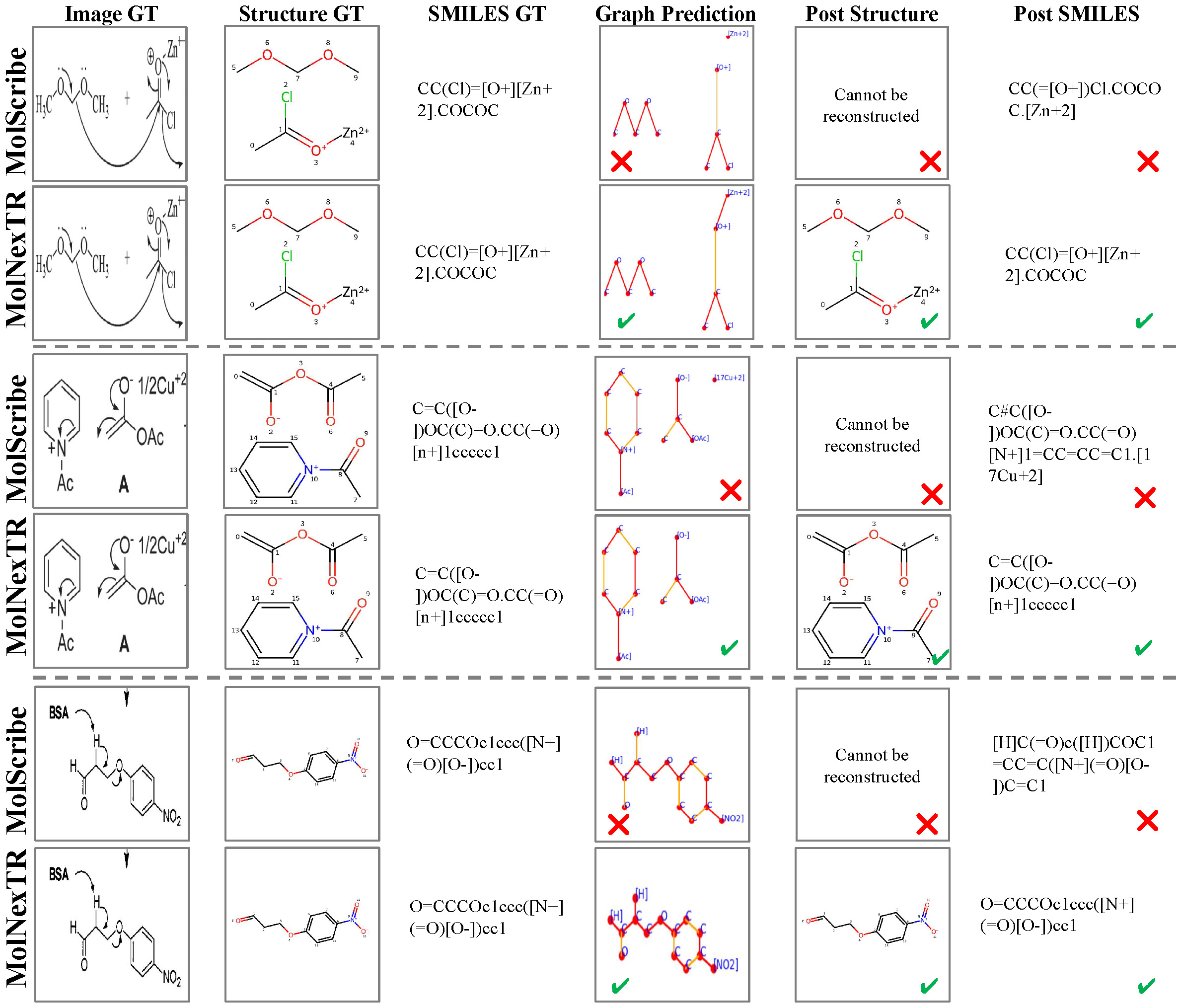}
\caption{Visualization of our model on some molecules in mechanism images compared to MolScribe.}                          
\label{fig:vs4}
\end{figure*}
In Fig.~\ref{fig:vs1}, we show some visualizations of the ACS dataset compared to the second-best method MolScribe. The experimental results show that the proposed MolNexTR method can accurately predict molecular structure from a variety of molecular images with various styles, especially those with contamination or complex abbreviations compared to MolScribe. 
The results demonstrate the importance of our series of data augmentation algorithms and abbreviation expansion and self-correction modules.

\subsubsection{Generalization of MolNexTR}
To further verify the generalization of our model, we also compared the model with the second-best method on some out-domain real hand-drawn molecules and molecules in mechanism images. 
Fig.~\ref{fig:vs3} shows the visualizations of some hand-drawn molecule images compared to the second-best method. 
Although the pattern of hand-drawn molecules is very different from the pattern of common molecular images (more random structures, more ambiguous atomic symbols, etc.), and we did not include any hand-drawn molecules in training, the experimental results show that our model performs better at recognizing all the bonds and atomic symbols compared to the current methods.

Fig.~\ref{fig:vs4} shows the visualizations of some molecules in mechanism images compared to the second-best method. MolNexTR can better recognize arrow symbols present in such molecules and is not affected by them, while the other method often obfuscates additional arrow or atom symbols as the main part of the molecular and outputs wrong predictions.
These results prove that MolNexTR has stronger generalization compared to the current methods.

\subsubsection{Error Cases Analysis}
\begin{figure*}[t]
\centering
\includegraphics[width=1\textwidth]{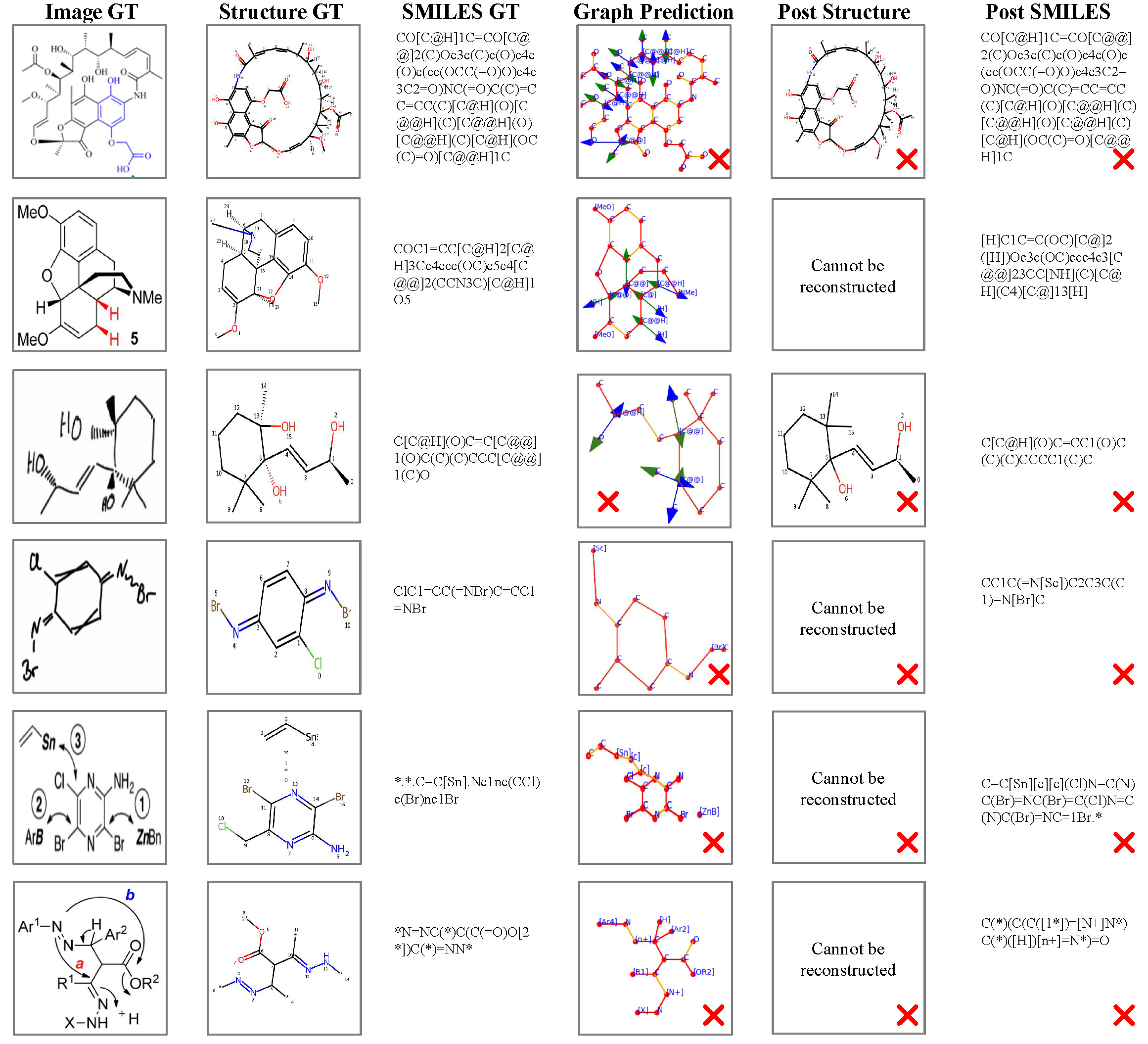}
\caption{Visualization of our model on some error cases.}                          
\label{fig:vs2}
\end{figure*}
We selected several cases with incorrect predictions for analysis. It is shown in Fig.~\ref{fig:vs2}.
We can see that MolNexTR tends to make wrong predictions for extremely complex molecules. The main reasons are these molecules have some rare structures, and the number of pixels assigned to each atom is low compared to other molecules. 
There are also cases where the drawing style causes the model to fail to recognize chirality, such as when dashed and solid wedges are replaced by broken lines. 
In addition, for some more complex hand-drawn molecules and molecules in mechanism images, MolNexTR still cannot guarantee the output of completely correct prediction results, though we did not include such molecules in training.
We hope that future work can continue to address these issues.
\section{Conclusions}
In this study, a new network architecture named MolNexTR is proposed for molecular image recognition. Compared with the advanced molecular image recognition architectures, MolNexTR achieves better recognition accuracy, and has better generalization and robustness.
The main contribution of this paper is the development of a novel two-stream encoder and advanced data augmentation algorithms, integrated with chemical knowledge, to significantly improve the model's feature extraction capabilities, robustness, and accuracy in predicting molecular structures. Experimental results on six public datasets verify the superiority of the MolNexTR method and the effectiveness of its components. 

One limitation is that there is still a lot of room for improvement in the accuracy of our model in Recognizing hand-drawn molecules and molecules in mechanism images as we mentioned before.
Another is that in real literature, R-group information in molecular images often appears in other places, such as texts or tables~\cite{wilary2021reactiondataextractor,AutomatedChemicalReactionExtractionfromScientificLiterature}. Future work may need to integrate this information to obtain a more complete molecular prediction.

\bmhead{List of abbreviations} OCSR: Optical Chemical Structure Recognition; MolNexTR: Molecular convNext-TRansformer; SMILES: Simplifed Molecular-Input Line-entry System; SMARTS: SMiles ARbitrary Target Specification; CNN: Convolutional Neural Network; ViT: Vision-Transformer; OCR: Optical Character Recognition; RNN: Recurrent Neural Networks; FFN: Feed-Forward Network; MHSA: Multi-Head Self-Attention; GPUs: Graphical Processing Units; Render Aug.: Render Augmentation; Image Aug.: Image Augmentation; Mol Aug.: Molecular Augmentation; Img Contam Alg.: Image Contamination Algorithm.

\section*{Declarations}
\bmhead{Availability of data and materials}
The source code of this article is available in \url{https://github.com/CYF2000127/MolNexTR}.
The datasets and the model checkpoint are available in \url{https://huggingface.co/datasets/CYF200127/MolNexTR/tree/main}.
\bmhead{Ethics approval and consent to participate}
Not applicable.
\bmhead{Consent for publication}
Not applicable.
\bmhead{Competing interests}
The authors declare no competing financial interest.

\bmhead{Funding}
This work was supported by HKUST (Project No. R9251, Z1269)

\bmhead{Authors' contributions}
YC wrote the main manuscript and developed the model. 
YC and CTL prepared all figures and tables collaboratively. 
CTL constructed the perturbed datasets for the model. YC, HG, and CTL designed all experiments collaboratively.
HC, HG, JS and YH supervised the work.
All authors contributed to the manuscript. All authors read and approved the final manuscript.
\bmhead{Acknowledgements}
We thank the Information Technology Services Center (ISTC) in HKUST for providing the HPC3 Cluster as our computational resource.

\bibliography{sn-bibliography}

\end{document}